\documentclass[10pt,twocolumn,letterpaper]{article}

\usepackage{iccv}      

\usepackage{times}
\usepackage{epsfig}
\usepackage{graphicx}
\usepackage{amsmath}
\usepackage{amssymb}

\usepackage{amsthm}
\theoremstyle{definition}

\usepackage{booktabs}
\usepackage{multirow}
\usepackage{xcolor}
\usepackage{enumitem}
\usepackage{colortbl} 

\definecolor{iccvblue}{rgb}{0.21,0.49,0.74}
\usepackage[pagebackref,breaklinks,colorlinks,allcolors=iccvblue]{hyperref}

\def\paperID{1544} 
\def\confName{ICCV}
\def\confYear{2025}

\title{Beyond Low-Rank Tuning: Model Prior-Guided Rank Allocation for Effective Transfer in Low-Data and Large-Gap Regimes.}

\author{Chuyan Zhang, Kefan Wang, Yun Gu*\\
School of Automation and Intelligent Sensing, \\
Institute of Medical Robotics,\\
Institute of Image Processing and Pattern Recognition,\\
Shanghai Jiao Tong University, Shanghai, China\\
{\tt\small yungu@ieee.org}
}

\begin{document}
\maketitle
\begin{abstract}
Low-Rank Adaptation (LoRA) has proven effective in reducing computational costs while maintaining performance comparable to fully fine-tuned foundation models across various tasks. However, its fixed low-rank structure restricts its adaptability in scenarios with substantial domain gaps, where higher ranks are often required to capture domain-specific complexities. Current adaptive LoRA methods attempt to overcome this limitation by dynamically expanding or selectively allocating ranks, but these approaches frequently depend on computationally intensive techniques such as iterative pruning, rank searches, or additional regularization. To address these challenges, we introduce Stable Rank-Guided Low-Rank Adaptation (SR-LoRA), a novel framework that utilizes the stable rank of pre-trained weight matrices as a natural prior for layer-wise rank allocation.  
By leveraging the stable rank, which reflects the intrinsic dimensionality of the weights, SR-LoRA enables a principled and efficient redistribution of ranks across layers, enhancing adaptability without incurring additional search costs. Empirical evaluations on few-shot tasks with significant domain gaps show that SR-LoRA consistently outperforms recent adaptive LoRA variants, achieving a superior trade-off between performance and efficiency. Our code is available at \url{https://github.com/EndoluminalSurgicalVision-IMR/SR-LoRA}.
\end{abstract}  
\section{Introduction}
Low-Rank Adaptation (LoRA) \cite{hulora} has become one of the most popular Parameter-Efficient Fine-Tuning (PEFT) methods, significantly reducing the number of trainable parameters while delivering competitive performance across many transfer learning tasks. This is achieved by freezing the pretrained model weights and injecting trainable low-rank matrices into specific layers as the low-rank approximation of the original model space. 
Although LoRA achieves significant parameter efficiency, recent studies have  disclosed that it struggles to outperform full fine-tuning (FFT), especially in more complicated downstream tasks~\cite{ding2023parameter, biderman2024lora, ren2024melora, jiang2024mora, chen2024quanta}. This discrepancy is attributed to the low-rank approximation mechanism, which is inadequate to fully capture the intricacies of complex downstream tasks or effectively learn and memorize new knowledge.

\begin{figure*}[!t]
    \centering
    \includegraphics[width=1\linewidth]{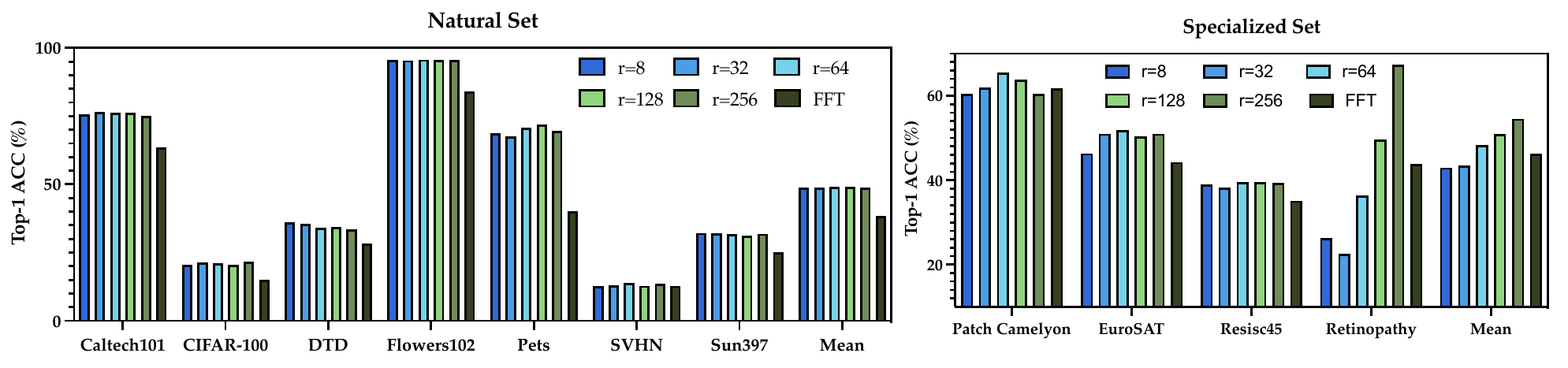}
    \caption{LoRA’s performance on different downstream tasks with a increasing rank.  For tasks on the VTAB natural set that are more similar to ImageNet-21k (i.e., with a smaller domain gap), the performance of LoRA is less affected by the rank and is significantly better than full fine-tuning (FFT). However, for tasks on the VTAB specialized set with a larger gap from ImageNet-21k, the average performance of LoRA improves progressively as the rank increases.}
    \label{fig:lora_rank}
    \end{figure*}

In this paper, we focus on a specific yet critical scenario where LoRA's limitations become particularly pronounced: few-shot learning with significant domain gaps. This limitation is empirically illustrated in Figure~\ref{fig:lora_rank}, which compares LoRA's performance across tasks from the VTAB benchmark~\cite{vtab}, categorized into the Natural Set and the Specialized Set. For tasks in the Natural Set, which exhibit smaller domain gaps relative to the pre-training dataset (i.e., ImageNet-21k), LoRA demonstrates robust performance across varying ranks and often outperforms FFT. This suggests that even a low rank is sufficient to capture the necessary adaptation for tasks with smaller domain shifts.  
\begin{figure}[!t]
    \centering
    \includegraphics[width=1\linewidth]{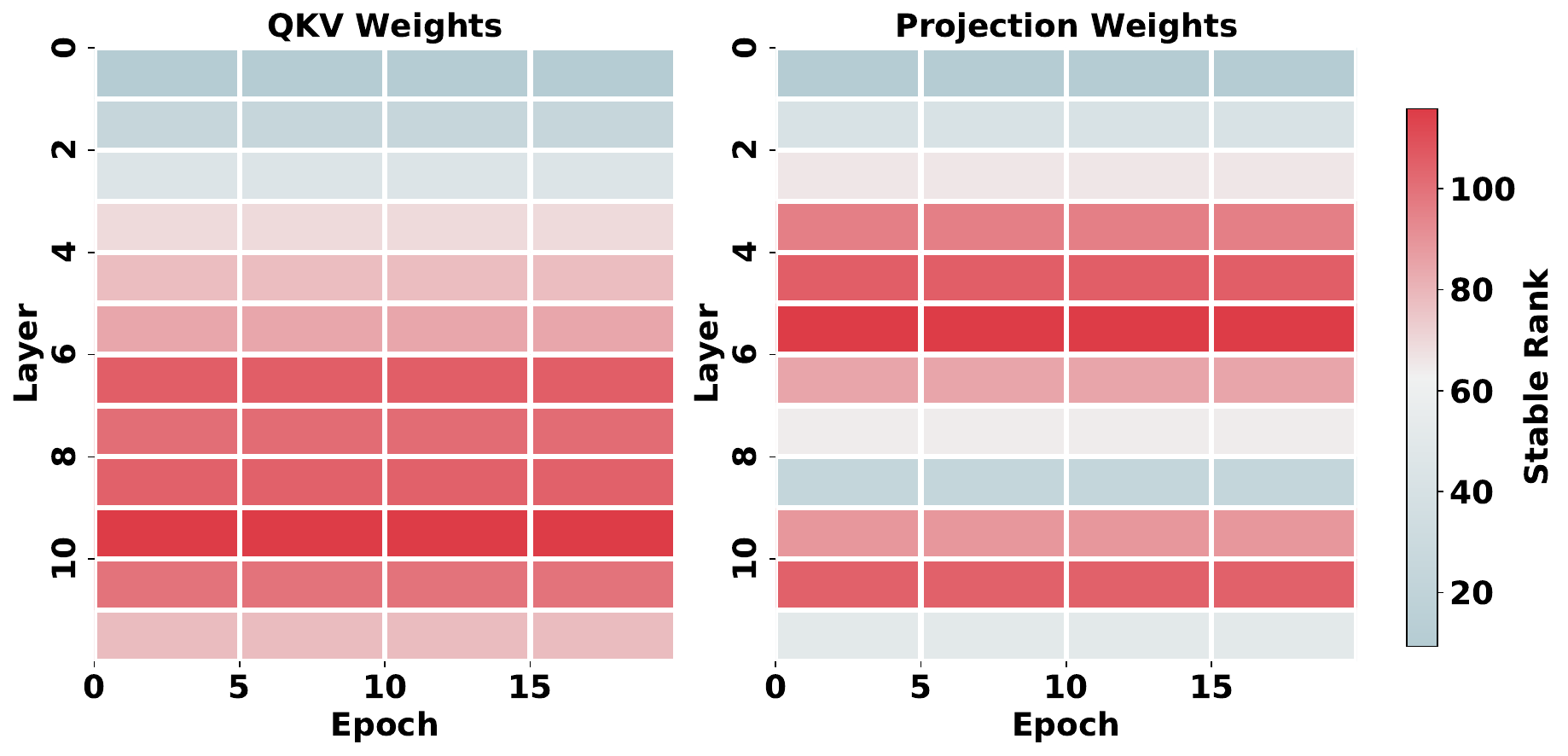}
    \caption{ The layer-wise stable rank of weights during the fine-tuning process of a pretrained model on a downstream task. }
    \label{fig:stable_rank_heatmap}
    \end{figure}
In contrast, for tasks in the Specialized Set, which involve larger domain gaps, the low-rank property becomes a bottleneck.
As the rank increases, LoRA's performance improves progressively, indicating that higher ranks are essential for handling the domain-specific complexities of these tasks. These findings highlight the importance of selecting an appropriate rank when dealing with more challenging transfer scenarios.

To address this issue, various approaches have been proposed to improve the adaptability of LoRA to downstream tasks. A line of work focuses on increasing the rank of LoRA while maintaining the same or fewer parameters. According to the sub-additivity property of the matrix rank, i.e., ${rank}(M_1 + M_2) \leq {rank}(M_1) + {rank}(M_2) $, the rank of incremental weight update can be expanded by ensembling multiple LoRA modules. Based on this principle, new forms of LoRA have been designed, including the merge-and-reinitialize procedure \cite{lialin2023relora,xia2024chainloraefficientfinetuning}, random masking~\cite{song2024increasing} or parallel stacking\cite{ren2024melora}. However, higher-rank tuning does not always lead to better results. Excessive LoRA ranks might lead to degradation in both performance and efficiency. Another line of work selectively allocates ranks to LoRA modules by assigning higher ranks to more critical layers or tasks and lower ranks to less important ones. Singular Value Decomposition (SVD) provides an effective tool to measure the contribution of different components in a matrix. AdaLoRA \cite{zhang2023adaloraadaptivebudgetallocation} incorporates orthogonality regularization for matrices $P$ and $Q$ to approximate SVD decomposition and then remove less important singular values using a novel importance scoring mechanism. 
Pissa~\cite{meng2024pissaprincipalsingularvalues} initializes LoRA modules using the principal singular components of the pretrained matrix. These components capture the most critical directions in the matrix, and aligning the initial weights with them helps to accelerate convergence and enhance performance. 

Although these LoRA variants have shown effectiveness, they often rely on complex hyperparameter tuning or optimization procedures, introducing additional computational overhead and complicating deployment in practical applications. This highlights the need for a more efficient approach to rank allocation that avoids excessive tuning complexity. 
In this paper, we propose that the stable rank of weights at each layer naturally reflects the inherent capacity of the pretrained model. Intuitively, layers with strong generalization capabilities can maintain effectiveness with a lower-rank approximation, as their encoded representations align well with diverse downstream tasks. Conversely, layers with limited generalization capacity require a higher rank to accommodate the greater degree of task-specific adaptation needed. 
The stable rank of parameters serves as a key indicator for a neural network's generalization behavior\cite{sanyalstable}. Specifically, a generalization bound is defined as $\mathcal{O}\left(\sqrt{\prod_{i} \|\mathbf{W}_i\|_2^2} \sum_{i=1}^d \text{srank}(\mathbf{W}_i)\right)$, which depends on both the Lipschitz constant upper-bound $\sqrt{\prod_{i} \|\mathbf{W}_i\|_2^2}$  (product of spectral norms) and the stable rank ($
\text{srank}$). A decrease in stable rank has been shown to reduce the Lipschitz constant, thereby implying better generalization performance. As illustrated in Figure~\ref{fig:stable_rank_heatmap}, the stable rank of a pretrained model tends to remain consistent across tuning epochs after pre-training, further supporting its reliability as a guiding metric.

Inspired by the intrinsic properties of stable rank, we propose a principled rank allocation strategy for LoRA, where the rank of each LoRA module is set as the stable rank of the corresponding pretrained model parameter matrix.
This simple yet effective approach enables flexible rank distribution across layers, aligning with the model's inherent structure and specific adaptation requirements. Unlike existing methods, it avoids the need for complex pruning, rank searches, or additional regularization, making it both practical and scalable for real-world applications.

\section{Related works}
 
\begin{figure*}[!t]
    \centering
    \includegraphics[width=0.86\linewidth]{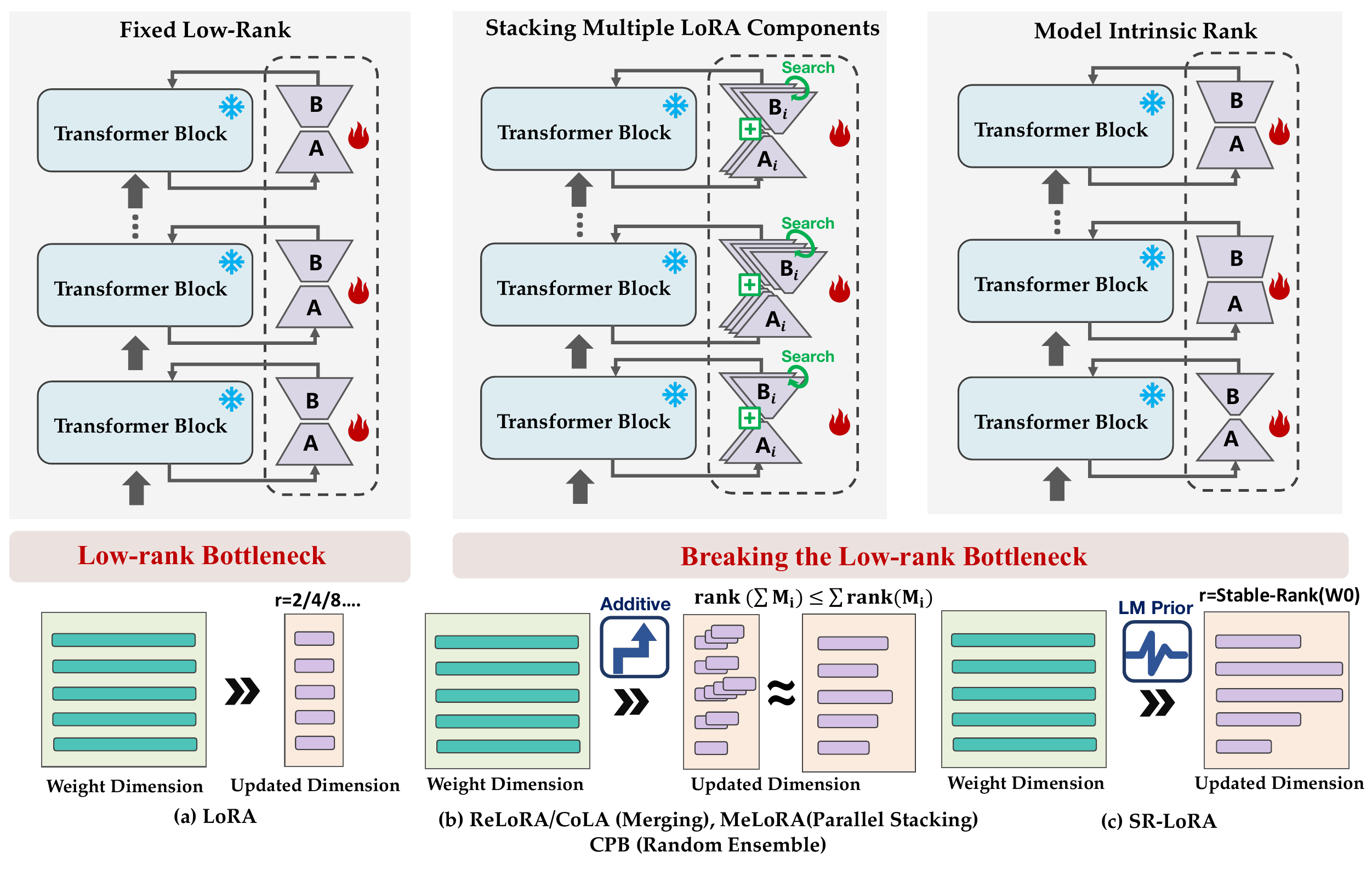}
    \caption{ An overview of the proposed SR-LoRA structure, compared to LoRA, DyLoRA and AdaLoRA. }
    \label{fig:lora_method}
    \end{figure*}

\textbf{Parameter-Efficient Fine-tuning (PEFT).} To reduce the high computational cost of fine-tuning large-scale foundation models, various parameter-efficient fine-tuning (PEFT) methods have been developed, which focus on updating only a small subset of (extra) model parameters. {VPT}~\cite{jia2022vpt} adds trainable visual prompts in the input space of foundation models. {Adapter}~\cite{houlsby2019adapter} introduces trainable modules within transformer layers, utilizing a down-sampling layer for dimensionality reduction, a non-linear activation function, and an up-sampling layer to restore dimensionality. {BitFit}~\cite{zaken2022bitfit} updates only the bias terms in the backbone. {SSF}~\cite{lian2022ssf} introduces trainable affine parameters to scale and shift the pretrained representations. {LoRA}~\cite{hulora} replaces full parameter updates with low-rank matrices to efficiently modify representations while maintaining dimensional transformations. 

\textbf{LoRA Variants.}  Recently, various LoRA variants have been proposed to overcome the low-rank bottleneck and improve adaptation performance.
ReLoRA \cite{lialin2023relora} and COLA \cite{xia2024chainloraefficientfinetuning} 
progressively merge the tuned LoRA modules into the frozen backbone, and then reinitialize and retrain them throughout the fine-tuning process.
However, the merge-and-reinitialize procedure does not always guarantee a rank increase, as overlap may occur among LoRA modules during fine-tuning. To address this issue, subsequent works stack multiple LoRA matrices to increase the rank.
MeLoRA~\cite{ren2024melora} proposes training a group of mini LoRA modules in parallel.
MoRA~\cite{jiang2024mora} replaces the low-rank matrices in LoRA with a square matrix of higher rank and integrates non-parameterized operators for input dimension reduction and output dimension expansion. CPB~\cite{song2024increasing} applies random masks within LoRA modules to aggregate a set of different weight matrices.
Beyond constructing new forms of LoRA matrices, some studies have explored adaptive allocation of ranks for LoRA modules from diverse perspectives, such as singular value decomposition (AdaLoRA~\cite{zhang2023adaloraadaptivebudgetallocation}, SaLoRA \cite{hu2023structure}and Pissa~\cite{meng2024pissaprincipalsingularvalues}), rank sampling (DyLoRA~\cite{valipour2022dylora}) and layer-wise structure search (GLoRA~\cite{chavan2023glora} and GeLoRA~\cite{ed2024gelora}).
Most of these studies have been validated on large language models (LLMs), while their effectiveness on vision models, particularly in large-gap and few-shot data regimes, remains to be explored.


\section{Rank-Guided Adaptation}
\subsection{Background and Intuitions}

\textbf{Low-Rank Adaptation (LoRA).} LoRA \cite{hulora} assumes that parameter updates during fine-tuning can be effectively approximated by low-rank transformations. To leverage this observation, LoRA freezes the pretrained weight matrix $W_{\text{pretrained}} \in \mathbb{R}^{d \times k}$ and introduces a low-rank update $\Delta W$ to adapt the model. The adapted weight matrix can be expressed as:
\begin{equation}
W_{\text{finetuned}} = W_{\text{pretrained}} + \Delta W, \quad \Delta W = BA,
\end{equation}
where $A \in \mathbb{R}^{d \times r}$, $B \in \mathbb{R}^{r \times k}$, and the rank $r$ satisfies $r \ll \{d, k\}$. In practical implementations, $A$ is initialized with random Gaussian values $A_0 \sim \mathcal{N}(0, 1)$, and $B$ is initialized to zero ($B_0 = 0$). This ensures that at the beginning of the training process, the perturbation $\Delta W$ is zero, preserving the original pretrained weights. Compared to full-rank adaptation, LoRA significantly reduces the number of trainable parameters. Specifically, instead of optimizing $d \times k$ parameters, LoRA introduces only $d \times r + r \times k$ parameters, which are much fewer when $r$ is small. This reduction leads to lower memory consumption and fewer FLOPS during gradient computation, making it particularly efficient for adapting large-scale pretrained models. In standard LoRA, the rank $r$ is a hyperparameter that needs to be adjusted for each specific task. The search process for the optimal rank configuration can be time-consuming and resource-intensive.

\textbf{Singular Value Decomposition (SVD).}
The singular value decomposition (SVD) of $\mathbf{W}$ is expressed in terms of its singular values, left and right singular vectors.
The $i$-th singular value of $\mathbf{W}$ is denoted as $\sigma_i(\mathbf{W})$. Using these singular values, we can compute the 2-norm and the Frobenius norm of the matrix. Specifically, the 2-norm $\|\mathbf{W}\|_2$ is given by the largest singular value $\sigma_1$, and the Frobenius norm $\|\mathbf{W}\|_F$ is computed as $\sqrt{\sum_i \sigma_i^2}$.


\textbf{Stable Rank.}
The stable rank of a matrix $\mathbf{W}$ is given by the ratio of its Frobenius norm squared to its spectral norm:
\begin{equation}
    \operatorname{srank}({W}) = \frac{\|\mathbf{W}\|_F^2}{\|\mathbf{W}\|_2^2} = \frac{\sum_{i=1}^{\operatorname{rank}(W)} \sigma_i^2(\mathbf{W})}{\sigma_1^2(\mathbf{W})},
\end{equation}
where $\operatorname{rank}(W)$ is the rank of the matrix
. Based on the previous research\cite{sanyalstable}, stable rank possesses the following properties:
\begin{enumerate}[nosep, left=0pt, label=\arabic*.]
\item  Stable rank is a smoothed version of the rank since it is more robust to small changes in the matrix.
\item  Stable rank is the lower bound for the matrix rank, i.e., $\operatorname{srank}(\mathbf{W}) = \frac{\sum_{i=1}^{\operatorname{rank}} \sigma_i^2(\mathbf{W})}{\sigma_1^2(\mathbf{W})} \leq \operatorname{rank}(W)$.
 \item The stable rank remains unchanged under scaling, meaning that for any \( \eta \in \mathbb{R} \setminus \{0\} \), it holds that \( \operatorname{srank}(\mathbf{W}) = \operatorname{srank}\left(\frac{\mathbf{W}}{\eta}\right) \).
\item Stable rank directly affects the generalization behaviour as increasing the stable rank directly decreases the lower bound of the noise sensitivity.
\end{enumerate}

\subsection{SR-LoRA: Allocating Rank with Model Prior}
Based on the above analysis, we present a search-free approach to improve the adaptation capability of low-rank updating by leveraging model prior. 
During the pretraining stage of a foundation model on a large-scale dataset, the stable rank of the parameters will gradually converge (see Figure~\ref{fig:stable_rank_heatmap}). When adapting the pretrained model to a few-shot downstream task, the learning rate is generally much smaller than during pretraining to avoid overfitting, and the stable rank of the model parameters remains largely unchanged. Consequently, the stable rank of the parameters in a pretrained model directly indicates the effective dimensionality of the parameter space. 

{Following \cite{hulora}, we only apply LoRA to the query, value and output projections (i.e., $W_q$, $W_v$ and $W_o$) in the Multi-head Attentions.} For parameter updating, we focus on the following subset of pretrained parameters, while all other parameters remain frozen:
\begin{equation}
   W_{pretrained} = \{W_{m,0}^{(l)}\} \\, m \in \{q, v, o\},  l \in [1, L]
\end{equation}
where $L$ is the layer number of the model. 
Intuitively, when fine-tuning starting from the pretrained model space, we can reduce the number of trainable parameters by directly using the low-rank adapters corresponding to its stable rank to approximate the original pretrained model. Specifically, we allocate the rank of each LoRA module with the corresponding stable rank of pretrained weights:
\begin{equation}
\Delta W = \{
   B_m^{(l)}A_m^{(l)}\mid r_m^{(l)}=\operatorname{srank}\{W_{m,0}^{(l)}\} \}
\end{equation}\\
\textbf{Stable Rank as a Lower Bound.} Based on the properties of stable rank, the rank of the introduced adaptation module $\Delta W$ serves as a lower bound for the rank of the pretrained model's parameter space:
\begin{equation}
\operatorname{rank}(\Delta W) = \operatorname{srank}(W_{pretrained}) \leq \operatorname{rank}(W_{pretrained})
\end{equation}
Through this approach, we identify the optimal low-rank estimation for each weight module. This avoids the negative effects of setting a fixed, overly small rank value (e.g., 8 in Figure~\ref{fig:lora_method}~(a)) that might disrupt the pretrained model's structure, as well as the extra overhead associated with heuristically searching for a better rank (Figure~\ref{fig:lora_method}~(b)).\\
\begin{figure}[!t]
    \centering
    \includegraphics[width=0.8\linewidth]{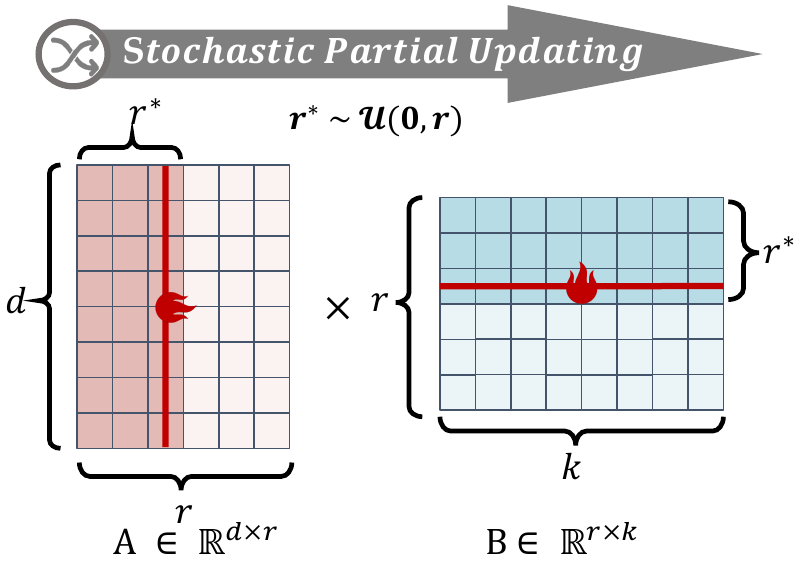}
    \caption{ Diagram of the stochastic partial updating strategy in the lightweight SR-LORA. }
    \label{fig:random_sampling}
    \end{figure}
\textbf{Lightweight SR-LORA.} The rank assigned by our method is smaller than the original parameter dimensions but could be larger than the commonly used small hyperparameter values. 
To reduce the number of trainable parameters while maintaining an effective dimensionality of the space, we adopt a \textbf{stochastic partial updating (SPU)} strategy.  
As illustrated in Figure~\ref{fig:random_sampling},  a value $r_s$ is randomly sampled from the range $[0, r]$ for a pair of $A$ and $B$ at each iteration. Only the first $r_s$ columns/rows of $A$ and $B$ participate in the forward pass, with parameter updates restricted to this selected column/row. Over multiple iterations, the full low-rank parameter space is learned progressively.

\begin{table*}[ht!]
\renewcommand{\arraystretch}{1} 
\centering
\caption{Performance of different PEFT methods on MedFM datasets for 1-5-10 shots.}
\resizebox{0.78\textwidth}{!}{
\begin{tabular}{llcccccccccc}
\toprule
\multirow{2}{*}{\footnotesize{\textbf{PEFT}}} & \multirow{2}{*}{\footnotesize{\textbf{n-shot}}} & \multicolumn{2}{c}{\footnotesize{\textbf{ChestDR}}} & \multicolumn{2}{c}{\textbf{\footnotesize{ColonPath}}} & \multicolumn{2}{c}{\footnotesize{\textbf{Endo}}} & \multirow{2}{*}{\footnotesize{\textbf{ALL}}} & \multirow{2}{*}{\footnotesize{\textbf{{Mean}}}} & \multirow{2}{*}{\footnotesize{\textbf{AUC}}} & \multirow{2}{*}{\footnotesize{\textbf{{Mean}}}} \\
\cmidrule(lr){3-4} \cmidrule(lr){5-6} \cmidrule(lr){7-8}
 & & \footnotesize{\textbf{mAP}} & \footnotesize{\textbf{AUC}} & \footnotesize{\textbf{ACC}} & \footnotesize{\textbf{AUC}} & \footnotesize{\textbf{mAP}} & \footnotesize{\textbf{AUC}} & & & & \\
\midrule
\multirow{3}{*}{Full-FT} 
& 1-shot & 13.93 & 58.65 & 82.73 & 83.77 & 18.22 & 59.50 & 52.80 &  & 67.31 &\\
& 5-shot & 14.78 & 64.85 & 79.33 & 89.02 & 18.47 & 65.44 & 55.32 & \cellcolor{gray!15}{56.26} & 73.10 & \cellcolor{gray!15}{72.32}\\
& 10-shot & 14.96 & 63.28 & 94.03 & 98.22 & 25.29 & 68.13 & 60.65 &  & 76.54 & \\
\midrule
\multirow{3}{*}{LP} 
& 1-shot & 12.27 & 54.78 & 77.57 & 80.63 & 17.84 & 57.38 & 50.08 &  & 64.26 & \\
& 5-shot & 15.76 & 63.37 & 79.90 & 89.51 & 19.13 & 63.08 & 55.13 & \cellcolor{gray!15}{55.28} & 71.99 &\cellcolor{gray!15}{71.42}\\
& 10-shot & 16.58 & 67.64 & 91.32 & 97.12 & 21.76 & 69.31 & 60.62 &  & 78.02 & \\
\midrule
\multirow{3}{*}{VPT~\cite{jia2022vpt}} 
& 1-shot & 12.83 & 56.72 & 76.14 & 80.76 & 19.77 & 61.42 & 51.27 &  & 66.30 & \\
& 5-shot & 15.41 & 63.47 & 88.36 & 95.92 & 18.07 & 63.27 & 57.42 & \cellcolor{gray!15}{55.62} & 74.22 &\cellcolor{gray!15}{71.92} \\
& 10-shot & 11.92 & 59.80 & 89.41 & 96.02 & 22.00 & 69.88 & 58.17 &  & 75.23 & \\
\midrule
\multirow{3}{*}{Adapter~\cite{houlsby2019adapter}} 
& 1-shot & 12.53 & 57.18 & 85.17 & 89.47 & 19.21 & 59.31 & 53.81 &  & 68.65 &  \\
& 5-shot & 11.01 & 57.95 & 85.63 & 97.04 & 21.66 & 65.22 & 56.42 & \cellcolor{gray!15}{57.07} & 73.40 & \cellcolor{gray!15}{72.98}\\
& 10-shot & 12.46 & 59.63 & 93.94 & 98.49 & 28.86 & 72.55 & 60.99 &  & 76.89 & \\
\midrule
\multirow{3}{*}{Bitfit~\cite{zaken2022bitfit}} 
& 1-shot & 10.77 & 51.39 & 67.03 & 76.62 & 18.11 & 64.13 & 48.01 & & 64.05 &  \\
& 5-shot & 14.10 & 61.14 & 89.07 & 96.53 & 21.18 & 64.26 & 57.71 & \cellcolor{gray!15}{56.01} & 73.98 & \cellcolor{gray!15}{72.29}\\
& 10-shot & 16.22 & 65.79 & 91.07 & 97.10 & 29.46 & 73.68 & \textbf{62.32} &  & 78.86 & \\
\midrule
\multirow{3}{*}{SSF~\cite{lian2022ssf}} 
& 1-shot & 13.53 & 56.38 & 80.99 & 85.49 & 21.20 & 63.75 & 53.56 &  & 68.54 & \\
& 5-shot & 16.07 & 66.44 & 86.66 & 94.22 & 21.26 & 63.58 & 58.04 & \cellcolor{gray!15}{57.33} & 74.75 & \cellcolor{gray!15}{73.42}\\
& 10-shot & 17.98 & 67.52 & 87.78 & 95.96 & 25.71 & 67.45 & 60.40 &  & 76.98 & \\
\midrule
\multirow{3}{*}{LoRA~\cite{hulora}} 
& 1-shot & 9.87 & 51.85 & 69.02 & 76.71 & 19.20 & 63.72 & 48.40 &  & 64.09 & \\
& 5-shot & 13.38 & 61.73 & 87.97 & 95.18 & 20.34 & 62.64 & 56.87 & \cellcolor{gray!15}{55.42} & 73.18 & \cellcolor{gray!15}{71.76}\\
& 10-shot & 16.83 & 67.07 & 88.73 & 95.78 & 26.46 & 71.13 & 61.00 &  & 77.99 & \\
\midrule
\multirow{3}{*}{SR-LoRA} 
& 1-shot & 13.33 & 57.70 & 81.63 & 86.69 & 22.93 & 73.01 & \textbf{55.88} & & \textbf{72.47} &  \\
& 5-shot & 16.52 & 67.39 & 88.52 & 96.10 & 19.88 & 65.12 & \textbf{58.54} & \cellcolor{gray!15}{\textbf{58.76}} & \textbf{76.20} & \cellcolor{gray!15}{\textbf{75.89}} \\
& 10-shot & 18.56 & 67.94 & 92.10 & 97.46 & 23.40 & 71.63 & 61.85 &  & \textbf{79.01} & \\
\bottomrule
\end{tabular}}\label{tab:medfm_peft}
\end{table*}
\section{Experiments Settings}
\subsection{Datasets}
\textbf{MedFM:} MedFM \cite{medfm}, a NeurIPS 2023 challenge, provides a comprehensive benchmark to evaluate the adaptation performance of foundation models on few-shot medical imaging tasks. The challenge focuses on foundation models pretrained on natural images and includes three publicly available tasks for evaluation: ChestDR, ColonPath, and Endo.
The ChestDR task involves the diagnosis of chest X-ray images across 19 diseases, formulated as a 19-way few-shot multi-label classification problem, and includes 2,140 training images, 2,708 validation images, and 2,626 test images. The ColonPath task classifies pathological images to determine the presence of tumors, framed as a binary classification problem, with 5,654 training images, 4,355 validation images, and 10,009 test images. The Endo task focuses on diagnosing colonoscopy images to identify three types of abnormalities and tumors, defined as a 3-way multi-label classification problem, and contains 1,810 training images, 2,055 validation images, and 2,199 test images.

\textbf{VTAB:} The VTAB-1k \cite{vtab} benchmark comprises 19 tasks spanning diverse domains: (1) natural images captured with standard cameras, (2) specialized images from non-standard imaging systems such as remote sensing and medical equipment, and (3) structured images generated from simulated environments. In this paper, we adopt a few-shot setup, where 1-shot training and validation datasets are randomly selected from the available 1,000 training samples for each task, while the test set remains consistent with the original VTAB-1k benchmark. Our comparison experiments primarily focus on specialized images, which exhibit larger domain gaps with pre-training data.


\begin{table}[!t]
\centering
\caption{Performance of different PEFT methods on VTAB-Specialized datasets. \texttt{LoRA-r*} refers to the specified rank in the LoRA method.}
\resizebox{0.4\textwidth}{!}{
\begin{tabular}{lccccc}
\toprule
\multirow{2}{*}{\textbf{Method}} & \multicolumn{5}{c}{\textbf{VTAB Specialized Datasets}} \\
\cmidrule(lr){2-5} 
 & \footnotesize{\rotatebox{90}{\textbf{Camelyon}}} & \footnotesize{\rotatebox{90}{\textbf{EuroSAT}}} & \footnotesize{\rotatebox{90}{\textbf{Resisc45}}} & \footnotesize{\rotatebox{90}{\textbf{Retinopathy}}} & \footnotesize{\rotatebox{90}{\textbf{Mean}}} \\
\midrule
Full-FT  & 61.75 & 44.30 & 35.17 & 43.88 & \cellcolor{gray!15}{46.27} \\
LP  & 55.43 & 40.71 & 34.91 & 46.44 & \cellcolor{gray!15}{40.70} \\
VPT~\cite{jia2022vpt} & 56.27 & 45.85 & 36.11 & 34.07 & \cellcolor{gray!15}{43.08} \\
Adapter~\cite{houlsby2019adapter}  & 62.97 & 48.86 & 35.24 & 54.00 & \cellcolor{gray!15}{50.27} \\
Bitfit~\cite{zaken2022bitfit} & 61.89 & 47.91 & \textbf{38.61} & 45.13 & \cellcolor{gray!15}{48.38} \\
SSF~\cite{lian2022ssf}  & 59.90 & 44.29 & 38.29 & 33.53 & \cellcolor{gray!15}{44.00} \\
\midrule
LoRA-r8~\cite{hulora}  & 60.40 & 46.36 & 38.95 & 26.25 & \cellcolor{gray!15}{42.99} \\
LoRA-r256~\cite{hulora} & 60.47 & 50.06 & 39.44 & 67.42 & \cellcolor{gray!15}{54.35} \\
MeLoRA~\cite{ren2024melora}  & 55.62 & 47.10 & 38.98 & 35.52 & \cellcolor{gray!15}{44.31} \\
GLoRA~\cite{chavan2023glora} & 58.26 & 34.08 & 22.57 & 26.34 & \cellcolor{gray!15}{35.31} \\
Pissa~\cite{meng2024pissaprincipalsingularvalues} & 57.69	&46.40	&\textbf{39.52}	&47.63	&\cellcolor{gray!15}{47.81} \\
CPB~\cite{song2024increasing} & 61.47 & 49.84 & 36.46 & 40.33 & \cellcolor{gray!15}{47.03} \\
MoRA~\cite{jiang2024mora}& 62.94 & \textbf{50.82} & 37.80 & 26.82 & \cellcolor{gray!15}{44.60} \\
DyLoRA~\cite{valipour2022dylora}  & 61.77 & 46.88 & 39.06 & 27.91 & \cellcolor{gray!15}{43.91} \\
\midrule
SR-LoRA & \textbf{64.22} & 49.70 & 38.01 & \textbf{73.60} & \cellcolor{gray!15}{\textbf{56.38}} \\
\bottomrule
\end{tabular}}\label{tab:vtab_peft}
\end{table}

\begin{table*}[ht!]
\renewcommand{\arraystretch}{1} 
\centering
\caption{Performance of different LoRA variants on MedFM datasets for 1-5-10 shots.\texttt{LoRA-r*} refers to the specified rank value used in the LoRA method.}
\resizebox{0.78\textwidth}{!}{
\begin{tabular}{llcccccccccc}
\toprule
\multirow{2}{*}{\footnotesize{\textbf{PEFT Method}}} & \multirow{2}{*}{\footnotesize{\textbf{n-shot}}} & \multicolumn{2}{c}{\footnotesize{\textbf{ChestDR}}} & \multicolumn{2}{c}{\textbf{\footnotesize{ColonPath}}} & \multicolumn{2}{c}{\footnotesize{\textbf{Endo}}} & \multirow{2}{*}{\footnotesize{\textbf{ALL}}} & \multirow{2}{*}{\footnotesize{\textbf{{Mean}}}} & \multirow{2}{*}{\footnotesize{\textbf{AUC \%}}} & \multirow{2}{*}{\footnotesize{\textbf{{Mean}}}} \\
\cmidrule(lr){3-4} \cmidrule(lr){5-6} \cmidrule(lr){7-8}
 & & \footnotesize{\textbf{mAP}} & \footnotesize{\textbf{AUC}} & \footnotesize{\textbf{ACC}} & \footnotesize{\textbf{AUC}} & \footnotesize{\textbf{mAP}} & \footnotesize{\textbf{AUC}} & & & & \\
\midrule
\multirow{3}{*}{LoRA-r8~\cite{hulora}} 
& 1-shot & 9.87 & 51.85 & 69.02 & 76.71 & 19.20 & 63.72 & 48.40 &  & 64.09 & \\
& 5-shot & 13.38 & 61.73 & 87.97 & 95.18 & 20.34 & 62.64 & 56.87 & \cellcolor{gray!15}{55.42} & 73.18 & \cellcolor{gray!15}{71.76}\\
& 10-shot & 16.83 & 67.07 & 88.73 & 95.78 & 26.46 & 71.13 & 61.00 &  & 77.99 & \\
\midrule
\multirow{3}{*}{LoRA-r256~\cite{hulora}} & 1-shot & 13.26 & 57.31 & 81.56 & 83.26 & 18.77 & 56.45 & 51.77 &  & 65.67 & \\& 5-shot & 15.88 & 65.19 & 87.81 & 96.14 & 19.66 & 64.84 & 58.25 & \cellcolor{gray!15}{57.01} & 75.39 & \cellcolor{gray!15}{72.86} \\& 10-shot & 16.83 & 66.02 & 91.66 & 97.44 & 25.12 & 69.06 & 61.02 &  & 77.51 & \\
\midrule
\multirow{3}{*}{MeLoRA~\cite{ren2024melora}} 
& 1-shot & 13.59 & 56.78 & 74.05 & 80.33 & 19.48 & 68.35 & 52.10 &  & 68.49 & \\
& 5-shot & 15.99 & 65.38 & 82.73 & 96.97 & 20.88 & 64.67 & 57.77 & \cellcolor{gray!15}{56.73} & 75.67 & \cellcolor{gray!15}{73.94} \\
& 10-shot & 17.38 & 66.47 & 86.75 & 95.70 & 24.83 & 70.77 & 60.32 &  & 77.65 & \\
\midrule
\multirow{3}{*}{GLoRA~\cite{chavan2023glora}} 
& 1-shot & 10.30 & 53.76 & 64.94 & 72.23 & 15.89 & 57.14 & 45.71 &  & 61.04 & \\
& 5-shot & 12.86 & 61.08 & 82.92 & 94.11 & 18.52 & 57.73 & 54.54 & \cellcolor{gray!15}{53.72} & 70.97 & \cellcolor{gray!15}{69.79} \\
& 10-shot & 14.95 & 64.03 & 91.41 & 97.32 & 27.14 & 70.68 & 60.92 &  & 77.34 & \\
\midrule
\multirow{3}{*}{Pissa~\cite{meng2024pissaprincipalsingularvalues}} 
& 1-shot & 13.38 & 56.58 & 78.14 & 80.87 & 18.37 & 60.44 & 51.30 &  & 65.96 & \\
& 5-shot & 15.18 & 64.26 & 89.80 & 97.13 & 21.99 & 65.56 & \textbf{58.99} & \cellcolor{gray!15}{56.96} & 75.65 & \cellcolor{gray!15}{72.94} \\
& 10-shot & 17.25 & 66.87 & 91.48 & 96.89 & 23.19 & 67.89 & 60.60 &  & 77.22 & \\
\midrule
\multirow{3}{*}{CPB~\cite{song2024increasing}} 
& 1-shot & 11.56 & 56.28 & 79.84 & 87.54 & 17.12 & 58.32 & 51.78 &  & 67.38 & \\
& 5-shot & 11.01 & 57.95 & 85.63 & 97.04 & 21.66 & 65.22 & 56.42 & \cellcolor{gray!15}{56.39} & 73.40 & \cellcolor{gray!15}{72.56} \\
& 10-shot & 12.46 & 59.63 & 93.94 & 98.49 & 28.86 & 72.55 & 60.99 &  & 76.89 & \\
\midrule
\multirow{3}{*}{MoRA~\cite{jiang2024mora}} 
& 1-shot & 10.69 & 53.38 & 57.31 & 78.45 & 20.23 & 67.65 & 47.95 &  & 62.94 & \\
& 5-shot & 13.60 & 61.99 & 79.89 & 95.86 & 18.73 & 62.93 & 55.50 & \cellcolor{gray!15}{54.72} & 73.59 & \cellcolor{gray!15}{72.67} \\
& 10-shot & 16.60 & 67.83 & 93.04 & 97.77 & 20.93 & 68.16 & 60.72 &  & 77.92 & \\
\midrule
\multirow{3}{*}{DyLoRA~\cite{valipour2022dylora}} 
& 1-shot & 13.75 & 58.01 & 78.55 & 80.51 & 25.15 & 72.67 & 54.77 &  & 70.40 & \\
& 5-shot & 15.71 & 64.66 & 78.85 & 96.62 & 19.77 & 64.59 & 56.70 & \cellcolor{gray!15}{57.18} & 75.29& \cellcolor{gray!15}{74.84}\\
& 10-shot & 18.24 & 67.51 & 79.43 & 96.97 & 26.31 & 71.98 & 60.07 &  & 78.82 & \\
\midrule
\multirow{3}{*}{SR-LoRA} 
& 1-shot & 13.33 & 57.70 & 81.63 & 86.69 & 22.93 & 73.01 & \textbf{55.88} & & \textbf{72.47} &  \\
& 5-shot & 16.52 & 67.39 & 88.52 & 96.10 & 19.88 & 65.12 & {58.54} & \cellcolor{gray!15}{\textbf{58.76}} & \textbf{76.20} & \cellcolor{gray!15}{\textbf{75.89}} \\
& 10-shot & 18.56 & 67.94 & 92.10 & 97.46 & 23.40 & 71.63 & \textbf{61.85} &  & \textbf{79.01} & \\
\bottomrule
\end{tabular}}\label{tab:medfm_lora}
\end{table*}
\subsection{Implementation details}
\textbf{Settings:}
All experiments are conducted on the PyTorch deep learning platform using NVIDIA GeForce RTX 3090 GPUs (24 GB). We build our implementation on the public codebase provided by MedFM \cite{medfm}. For each downstream task, a trainable classifier is attached to the pretrained backbone. In \textbf{Full-FT}  (Full Fine-Tuning), all parameters are trainable, whereas in  \textbf{LP} (Linear Probing), only the classifier is updated. For all other PEFT methods, the backbone parameters remain frozen, with only the introduced adaptation modules being trained. We fine-tune each model using the AdamW optimizer with a cosine annealing learning rate schedule, setting the initial learning rate to 1e-3 and the weight decay to 5e-2.  Each run consists of 20 epochs with a batch size of 4, and the best model is selected based on validation set performance.

\textbf{Evaluation Metrics:} We use task-specific evaluation metrics to assess performance. Following MedFM \cite{medfm}, we report mean Average Precision (mAP) and AUC for the ChestDR and Endo tasks, while for the ColonPath task, we use mean Accuracy (ACC) and AUC. For VTAB, we follow \cite{jia2022vpt} and adopt mean Accuracy (ACC) for each task.

\begin{figure}[!tb]
    \centering
    \includegraphics[width=0.9\linewidth]{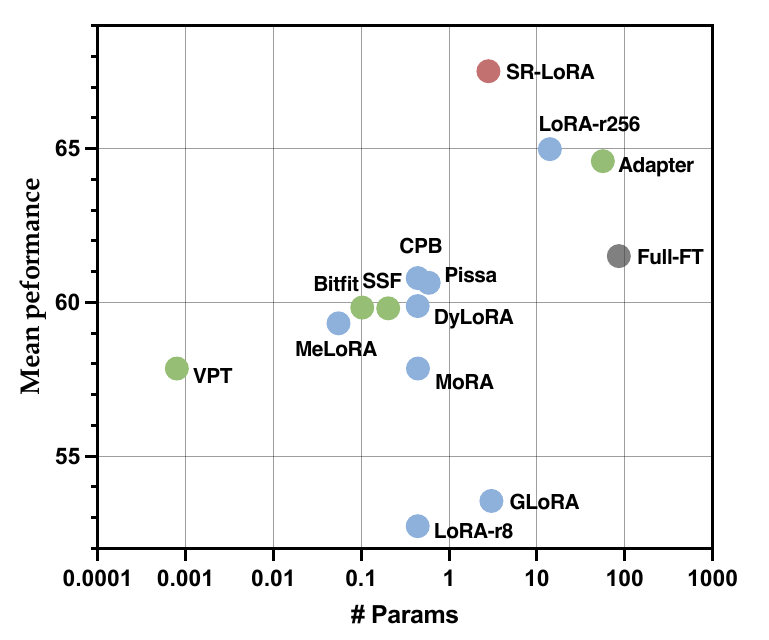}
    \caption{Comparison of parameter counts and average performance across MedFM-ChestDR, MedFM-ColonPath, MedFM-Endo, VTAB-Camelyon, and VTAB-Retinopathy datasets under the 1-shot setting. ``\# Params'' specifies the number of trainable parameters.}
    \label{fig:params}
    \end{figure}
\subsection{Results on MedFM and VTAB}
In this section, we utilize ViT-B16, a Vision Transformer model pretrained on ImageNet-21K following the standard supervised learning protocol, as the foundational model.

\textbf{Results on MedFM.} Table~\ref{tab:medfm_peft} presents the performance of various PEFT methods on the MedFM datasets under 1-shot, 5-shot, and 10-shot settings. Methods such as LP, VPT, Bitfit, and LoRA generally lag behind Full-FT in most cases. In contrast, SSF and Adapter perform better in 1-shot and 5-shot scenarios, benefiting from their more complex adaptation modules (the number of trainable parameters is shown in Figure~\ref{fig:params}). This highlights the importance of enhancing the model's adaptation capability in data-limited and challenging tasks. Overall, the proposed SR-LoRA outperforms traditional FFT and other PEFT methods, yielding the best performance across all domains in the MedFM dataset. In Table~\ref{tab:medfm_lora}, we compare SR-LoRA with various LoRA variants. First, we observe that simply increasing the LoRA rank to 256 (LoRA-256) significantly improves performance in 1-shot and 5-shot settings. By leveraging the model's prior knowledge, SR-LoRA adaptively allocates layer-wise ranks, further boosting performance. Additionally, DyLoRA, which employs a stochastic partial updating scheme, not only improves performance but also reduces computational costs, demonstrating that introducing randomness can enhance both efficiency and accuracy in few-shot learning tasks.

\textbf{Results on VTAB.} The 1-shot adaptation performance of different PEFT methods and LoRA variants on specialized datasets from the VTAB benchmark is shown in Table~\ref{tab:vtab_peft}. This evaluation underscores the challenges of adapting models pretrained on natural image datasets (ImageNet) to cross-modal tasks including medical (Camelyon and Retinopathy), satellite (EuroSAT) and remote sensing (Resisc45) imaging. The results indicate that performance on medical imaging datasets is more sensitive compared to satellite and remote sensing tasks.
This is expected, as medical imaging tasks require fine-grained detail analysis, whereas scene-based datasets focus on broader spatial patterns. 
Notably, the Retinopathy dataset involves grading Diabetic Retinopathy (DR) on a 0–4 scale, making it inherently more complex than object recognition (EuroSAT and Resisc45) or disease classification (Camelyon). The substantial gains achieved by high-rank tuning in LoRA-r256 and SR-LoRA align with previous studies~\cite{biderman2024lora,chen2024quanta}, suggesting that low-rank approximations may fail to handle complicated downstream tasks. While other methods struggle to consistently outperform FFT, SR-LoRA demonstrates significant improvements across all four tasks.

\textbf{Ablation study of rank allocation strategies.} We further conduct an ablation study of rank allocation strategies in Table~\ref{tab:ablation}. First, SPU performs comparably to the fixed-rank scheme while updating only $1/r$ parameters in LoRA modules per step, offering better computational efficiency. Second, blindly increasing the rank proves suboptimal compared to the proposed fine-grained allocation method based on the stable rank. Specifically, LoRA-r32 and SR-LoRA introduce the same number of trainable parameters, but SR-LoRA achieves significantly better performance.

\textbf{More architectures and pretraining paradigms.} In Table~\ref{tab::arch_pre}, we further evaluate SR-LoRA on ViT-L and Swin-Transformer with DINO and MAE pretraining schemes. As the network scale increases (ViT-B $\rightarrow$ ViT-L), the performance of FFT in the 1-shot setting deteriorates instead of improving. This indicates that an over-parameterized model tends to overfit when trained on limited data. In general, SR-LoRA achieves the best performance, demonstrating its robustness. 
\subsection{Analysis and Discussion}
\begin{figure}[!tb]
    \centering
    \includegraphics[width=0.9\linewidth]{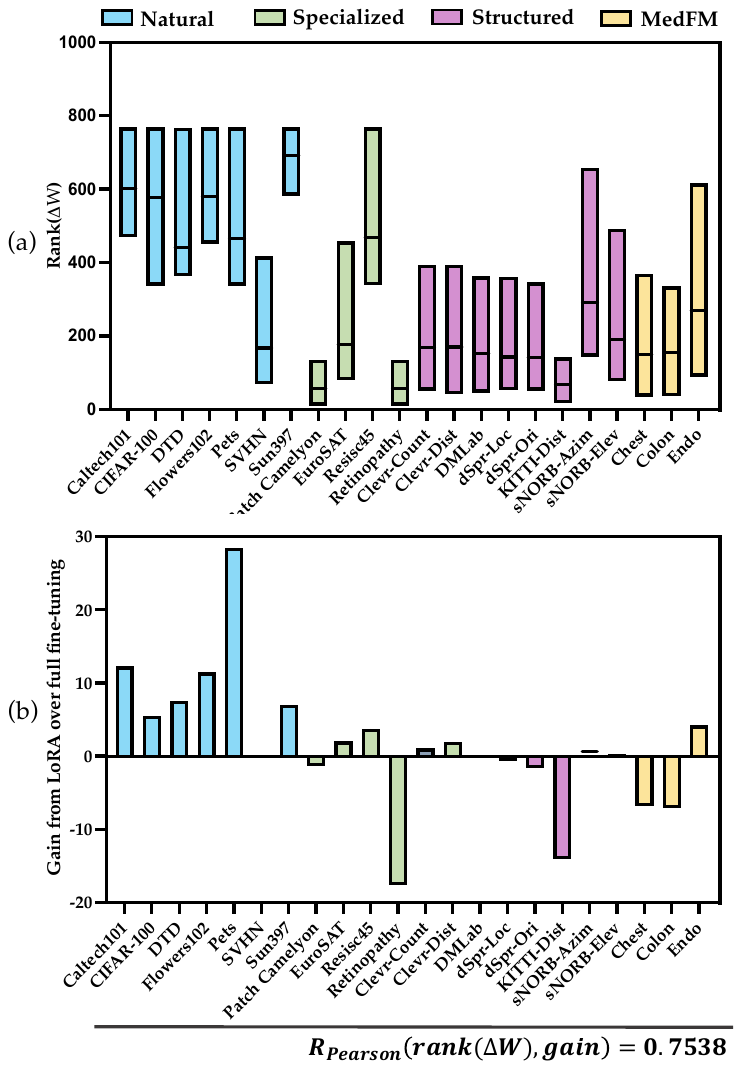}
    \caption{Dynamics of rank for pretrained ViT tuned on various downstream datasets. The x-axis denotes the downstream dataset name and the y-axes represent: (a) illustrates the rank of the tuned parameters across layers, and (b) shows the performance gain of LoRA tuning compared to full fine-tuning. We observe a strong positive correlation (Pearson correlation of 0.7538) between these two quantities represented on the y-axes.  }
    \label{fig:delta_w_rank}
    \end{figure}
\textbf{Updated parameter space.} As shown in Figure~\ref{fig:delta_w_rank}, the high rank of the adjusted parameter matrix via FFT in the natural set indicates that full fine-tuning introduces great complexity to the pretrained model, which may lead to overfitting on extremely limited data. In such cases, reducing the number of trainable parameters (e.g., using LoRA) can effectively improve downstream task performance. However, for other datasets, the model complexity is inherently low. Hence, expanding the model's parameter space should be prioritized, as applying low-rank estimation further restricts the model's capacity, resulting in suboptimal performance. 

\begin{figure}[!tb]
    \centering
    \includegraphics[width=1\linewidth]{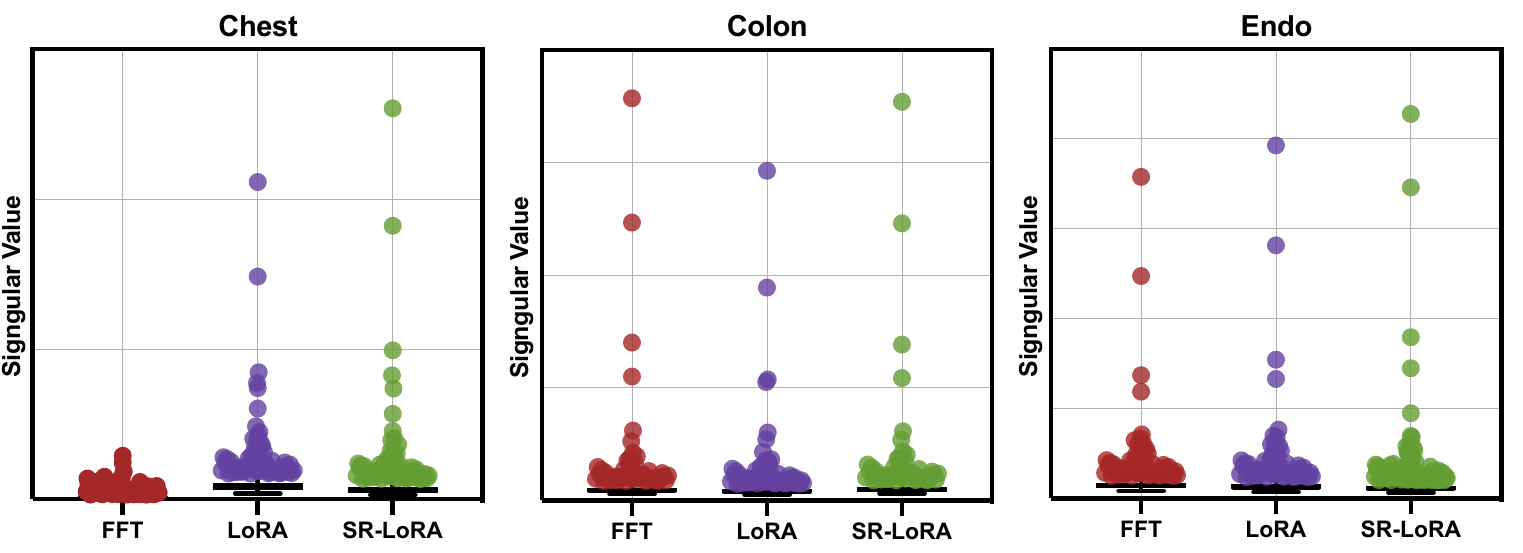}
    \caption{Distribution of Singular Values via feature SVD on MedFM datasets for different tuned models.}
    \label{fig:feature_svd}
    \end{figure}

\textbf{Feature space.} In addition to the parameter space, we extract the latent-space features before the classifier from the test set and examine the distribution of singular values via SVD. Previous work~\cite{pmlr-v97-chen19i} has demonstrated that the transferability of features is often concentrated in feature vectors associated with large singular values. As shown in Figure~\ref{fig:feature_svd}, SR-LoRA increases the number of large singular values, indicating enhanced transferability of the model.

\begin{table}[!th]
\centering
\caption{Ablation study of different rank allocation strategies in LoRA. \texttt{Fixed} means use the unifed rank for all layers, \texttt{SPU} refers to the Stochastic Partial Updating, and \texttt{SR-LoRA*} denotes SR-LoRA without SPU. `` Params\%'' specifies the ratio of trainable parameters to the total model parameters. Here, we report the average AUC (\%) over three MedFM datasets and the average Top-1 accuracy (\%) over four VTAB-Specialized datasets.}
\resizebox{0.42\textwidth}{!}{
\begin{tabular}{lccccc}
\toprule
\multicolumn{1}{c}{\multirow{2}{*}{\textbf{\small{Method}}}} & \multicolumn{3}{c}{\textbf{\small{MedFM}}}                        & \textbf{\small{VTAB-Spe}}       & \multirow{2}{*}{\textbf{\small{Params\%}}} \\ \cmidrule(r){2-4}\cmidrule(r){5-5}
\multicolumn{1}{c}{}                        & {1-shot}         & {5-shot}         & {10-shot}        & {1-shot}         &                               \\ \midrule
Fixed-r8                                     & 70.08          & 74.25          & 74.84          & 42.99          & \multirow{2}{*}{0.52\%}                          \\
SPU-r8          & 70.40          & 75.29          & 78.82          & 43.91          &                         \\\midrule
Fixed-r64                                  &    66.32       &      75.77    &     78.41     &      48.31  &             \multirow{2}{*}{4.13\%}         \\
SPU-r64          &  68.79        &   75.49      &  78.66        &       48.73       &                     \\\midrule
Fixed-r128                                  & 69.00          &       75.42   &     78.24     &    50.88    &  \multirow{2}{*}{8.25\%}                      \\
SPU-r128           &  68.91         &  75.74        &   78.06        &    49.47          &                         \\\midrule
Fixed-r256                                   & 65.67          & 75.39          & 77.51          & 54.35        &  \multirow{2}{*}{16.50\%}                        \\
SPU-r256           & 68.91          & 75.74          & 78.06          &        53.69        &                              \\\midrule
SR-LoRA$^*$                & 71.32         & 75.79          & 78.31          & \textbf{57.69} & \multirow{2}{*}{4.52\%}                          \\ 
SR-LoRA                                   & \textbf{72.47} & \textbf{76.20} & \textbf{79.01} & 56.38          &                 \\ \bottomrule
\end{tabular}}\label{tab:ablation}
\end{table}

\begin{table}[]
\caption{The average one-shot Top-1 accuracy (\%) over four VTAB-specialized datasets using various pretrained transformers with different PEFT methods. We note the total number of parameters for each backbone and the ratio of trainable parameters in PEFT methods to the total parameters.}
\centering
\resizebox{0.45\textwidth}{!}{
\begin{tabular}{ccccc}
\hline
Init.   & \multicolumn{4}{c}{ImageNet21k} \\ 
\hline
\multirow{2}{*}{Method}  & {ViT-L} & {Swin-T} & {Swin-S} & {Swin-B} \\ 
        & \small{303 M} & \small{28 M} & \small{49 M} & \small{87 M} \\ 
\hline
FFT         & 46.26  /\scriptsize{100\%}              & 34.16 /\scriptsize{100\%}      & 38.67  /\scriptsize{100\%}     & 38.45  /\scriptsize{100\%}     \\
VPT           & 42.15 /\scriptsize{0.00\%}        & 35.17 /\scriptsize{0.00\%}     & 46.86 /\scriptsize{0.00\%}      & 44.90 /\scriptsize{0.00\%}    \\
BitFit        & 47.37 /\scriptsize{0.09\%}          & 28.93  /\scriptsize{0.27\%}     & 34.41  /\scriptsize{0.31\%}    & 48.93  /\scriptsize{0.23\%}    \\
LoRA           & 42.88  /\scriptsize{0.39\%}         & 31.91 /\scriptsize{1.54\%}     & 30.95 /\scriptsize{1.77\%}   & 34.82 /\scriptsize{1.33\%}     \\
SR-LoRA  & \textbf{53.37} /\scriptsize{4.27\%} & \textbf{51.07} /\scriptsize{7.73\%}  & \textbf{52.92}  /\scriptsize{8.66\%} & \textbf{50.76} /\scriptsize{8.50\%} \\ \hline
Init.  & \multicolumn{2}{c}{DINO} & \multicolumn{2}{c}{MAE} \\ \hline
\multirow{2}{*}{Method}  & ViT-B       & ViT-L      & ViT-B      & ViT-L      \\
  & \small{85 M} & \small{303 M} & \small{85 M} & \small{303 M} \\ 
\hline
FFT          & 51.69 /\scriptsize{100\%}        & 32.35 /\scriptsize{100\%}       & 42.06 /\scriptsize{100\%}        & 28.02 /\scriptsize{100\%}       \\
VPT         & 41.65 /\scriptsize{0.00\%}      & 25.64 /\scriptsize{0.00\%}     & 30.47 /\scriptsize{0.00\%}     & 30.38 /\scriptsize{0.00\%}     \\
BitFit      & 42.88 /\scriptsize{0.12\%}         & 30.51 /\scriptsize{0.09\%}         & 41.64 /\scriptsize{0.12\%}         & 30.95 /\scriptsize{0.09\%}         \\
LoRA        & 45.47 /\scriptsize{0.52\%}       & 36.44 /\scriptsize{0.39\%}         & 42.24 /\scriptsize{0.39\%}        & 32.80   /\scriptsize{0.52\%}      \\
SR-LoRA &  \textbf{52.91} /\scriptsize{4.46\%}  & \textbf{43.59}/\scriptsize{4.15\%}  & \textbf{49.33} /\scriptsize{4.18\%}  & \textbf{46.10} /\scriptsize{4.11\%}  \\ \hline
\end{tabular}}\label{tab::arch_pre}
\end{table}

\section{Conclusion}
This paper addresses the limitations of LoRA in few-shot learning scenarios with significant domain gaps. Empirical evidence shows that the fixed low-rank approximation often struggles to capture the complex adaptations needed for tasks with significant domain gaps. To overcome this challenge, we propose a novel Stable Rank-based LoRA (SR-LoRA) method. SR-LoRA leverages the intrinsic properties of stable rank, which naturally reflects the generalization capacity of the pretrained model. By assigning the rank of each LoRA module to match the stable rank of the corresponding pretrained weight matrix, SR-LoRA enables flexible and adaptive rank distribution across layers. Unlike prior adaptive methods that rely on complex optimization or iterative pruning, our approach provides a straightforward and scalable solution for fine-tuning with LoRA. Experimental results prove that SR-LoRA not only improves the adaptability of LoRA but also maintains its simplicity and computational efficiency. Future work will explore the application of stable rank-based strategies to other PEFT methods and their extension to broader domains and tasks.

{
    \small
    \bibliographystyle{ieeenat_fullname}
    \bibliography{main}
}

\end{document}